\documentclass[journal, twoside]{IEEEtran}
\usepackage[english]{babel}

\usepackage{amsfonts}
\usepackage{amsmath}
\usepackage{amsthm}
\usepackage{caption}
\usepackage{graphicx, subfigure}
\usepackage[hidelinks]{hyperref}
\usepackage{algpseudocode}
\usepackage{algorithm}
\usepackage{float} 
\usepackage{booktabs}
\usepackage{multirow}
\usepackage{multicol}

\usepackage{lipsum}
\usepackage[utf8]{inputenc}
\usepackage[T1]{fontenc}

\usepackage{nccmath}
\usepackage{color}
\usepackage{cite}
\usepackage{scalerel}

\usepackage{yaacro}

\usepackage[colorinlistoftodos]{todonotes}

\newcommand{\etal}{\textit{et al.}}
\newcommand{\eg}{\textit{e.g.}}
\newcommand{\ie}{\textit{i.e.}}

\begin{acgroupdef}[list=acronimos]
    \acdef{CNPq}{Conselho Nacional de Desenvolvimento Científico e Tecnológico}
    \acdef{CAPES}{Coordenação de Aperfeiçoamento de Pessoal de Nível Superior}
    \acdef{FAPEMIG}{Fundação de Amparo à Pesquisa do Estado de Minas Gerais}
    \acdef{MRS}{Multi-Robot Systems}
\end{acgroupdef}

\newtheorem{definition}{Definition}  
\newtheorem{problem}{Problem}

\usepackage{hyperref} 
\begin{document}

\title{Communication Backbone Reconfiguration\\with Connectivity Maintenance}

\author{
Leonardo Santos, Caio C. G. Ribeiro, and Douglas G. Macharet
\thanks{This work was supported by CAPES/Brazil - Finance Code 001, CNPq/Brazil, and FAPEMIG/Brazil.}
\thanks{The authors are with the Computer Vision and Robotics Laboratory (VeRLab), Department of Computer Science, Universidade Federal de Minas Gerais, MG, Brazil. E-mails: {\tt\small leohmcs@ufmg.br, caioconti@ufmg.br, doug@dcc.ufmg.br}
}}

\maketitle

\pagestyle{empty}
\thispagestyle{empty}

\begin{abstract}
The exchange of information is key in applications that involve multiple agents, such as search and rescue, military operations, and disaster response.
In this work, we propose a simple and effective trajectory planning framework that tackles the design, deployment, and reconfiguration of a communication backbone by reframing the problem of networked multi-agent motion planning as a manipulator motion planning problem. 
Our approach works for backbones of variable configurations both in terms of the number of robots utilized and the distance limit between each robot. 
While research has been conducted on connection-restricted navigation for \ac{MRS} in the last years, the field of manipulators is arguably more developed both in theory and practice. 
Hence, our methodology facilitates practical applications built on top of widely available motion planning algorithms and frameworks for manipulators. 
\end{abstract}



\section{Introduction} \label{sec:introducao}

\ace{MRS} have been studied in a variety of applications, such as mapping \cite{banfi2018strategies}, target tracking~\cite{van2020mrs-tracking}, warehouse management \cite{ribeiro2022collaborative} and search and rescue \cite{drew2021intro-search-and-rescue}. A particularly compelling application of \ac{MRS} entails leveraging groups of mobile robots as relays to establish \textit{ad hoc} networks when communication infrastructure is not available. These systems allow a robot to collect information and then transmit it through a chain of relay robots to the base station. This capability is fundamental in disaster response scenarios~\cite{queralta2020collaborative}, where robots are used to collect and transmit information to human rescuers. We consider the scenario of a leader robot sending information to the next robot in line, which then passes it along to the subsequent robots until it reaches a base station.



\begin{figure}[t]
\centering
\includegraphics[width=.8\linewidth, trim={0cm 4cm 0cm 0cm},clip]{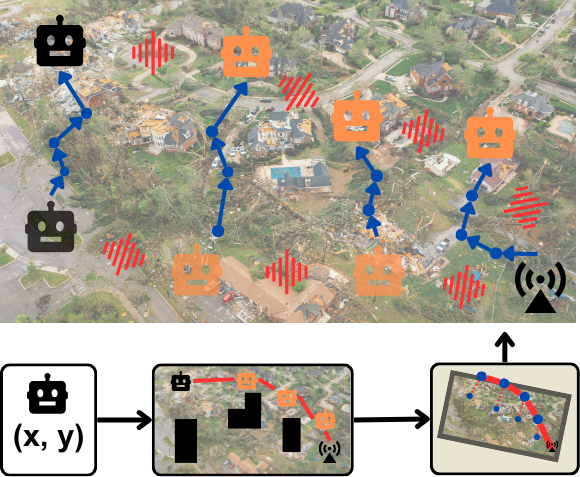}
\caption{Illustration of a backbone reconfiguration with a different number of robots. The initial configuration is at the bottom, and the new one is shown at the top. The leader robot is depicted in black, and relay robots in orange. Planned trajectories (blue arrows), are guaranteed by construction to keep the backbone connected throughout the trajectories execution.}
\label{fig:test}
\end{figure}

We address the problem of backbone reconfiguration with connectivity maintenance. This involves two key components: first, determining the endpoint locations of the relay robots to form a \emph{backbone} that connects a designated \emph{leader robot} to a \emph{base station}; second, navigating the robots to these positions to establish the computed backbone.
Given the high cost of robotic assets, we also optimize for the minimum number of robots required to complete the operation efficiently.
In scenarios featuring intermittent connectivity \cite{kantaros2019temporal}, each robot can individually compute its path, under the assumption that continuous connection is unnecessary during transit. In our work, we address the more complex case where sustained end-to-end connectivity is required at all times, focusing on line-of-sight connectivity. Therefore, the robots' trajectories must be planned collectively to ensure network continuity during movement, and not only in the final backbone (see Fig.~\ref{fig:test}).

We determine the robots' final positions using a visibility graph, post-processed to utilize the minimum number of relay robots.
Next, to maintain network connectivity during movement, we adopt the virtual structure concept~\cite{Lewis1997HighPrecision}, modeling the \ac{MRS} as a two-dimensional, linearly connected network. This approach allows us to apply any single-robot planning algorithm to efficiently plan coordinated paths for the entire robot team.
However, the virtual structure framework does not accommodate a variable number of robots or adjustable inter-robot distances. To address these limitations, we expand the state space to three dimensions, plan the trajectory in 3D, and then project the solution back into the 2D space.
The 3D structure maintains its shape while allowing adjustments to the number of robots and their distances in 2D. By treating the 3D structure as a robotic manipulator, we can apply inverse kinematics and manipulator planning algorithms to generate the trajectory in 3D, indirectly solving the 2D multi-robot navigation problem while maintaining line-of-sight connectivity.

In summary, the main contributions of this paper are:
\begin{enumerate}
    \item A complete framework to solve both the network configuration calculation (relay robots endpoint locations) and the networked \ac{MRS} path planning problem;
    \item A new modeling that allows to use standard single-robot planners in a linearly connected \ac{MRS}, with built-in guarantees for maintaining line-of-sight connectivity. 
\end{enumerate}

\section{Related Work} \label{sec:related_work}

The problem of planning robot trajectories while maintaining connectivity to a base station has been studied in both wired~\cite{Shapovalov2020Tangle} and wireless~\cite{hsieh2008maintaining} networks. Given the physical limitations of wired connections, such as entanglement, we focus on the wireless case.
Also, there are two main approaches to modeling wireless networks. One uses realistic communication models based on statistical methods~\cite{mostofi2011camp}, while the other assumes that two robots are connected if they are within a threshold distance and have line-of-sight~\cite{yang2010decentralized}. Although more conservative, the latter approach is simpler and only requires a workspace map, providing reliable results for our application.

Deploying networked robot systems becomes challenging when there are obstacles in the environment. In this scenario, Stump~\etal~\cite{stump2011visibility} propose a framework capable of deploying and redeploying a set of router robots using a polygonal map decomposition, considering intermittent connectivity. 
Santos~\etal~\cite{Santos2019Fast} propose a graph-based approach considering Steiner trees for deployment of a set of router robots, not considering reconfiguration for the case where clients' positions may change.
Wang~\etal~\cite{wang2015optmism} also tackle the problem of connecting a single robot to a base station via an end-to-end network of router robots, but they do not guarantee connectivity while the robots are moving to their final destinations.
Our work extends these approaches by ensuring continuous connectivity of the relay robots, during initial deployment and reconfiguration, specially while they are moving to new positions.

In order to guarantee connectivity during reconfiguration, we must solve a communication-restricted planning problem. In this context, Banfi~\etal~\cite{banfi2016asynchronous} propose planning algorithms for exploration in communication-restricted environments based on a line-of-sight model; however, they consider recurrent or periodic, not online connectivity. Online connectivity maintenance with a fixed base station has been addressed by Pei~\etal~\cite{pei2010coordinated}, where they propose a centralized algorithm that runs on the base station to solve a variant of the Minimum Steiner Tree Problem.
Global connectivity maintenance in \ac{MRS} has been successfully achieved by control frameworks proposed by Luo~\etal~\cite{luo2020behavior}. An advantage of our methodology over these works is the reduced complexity of implementation, leveraging widely available motion planning resources for manipulators, thus facilitating applications. 

In this work, we propose a centralized approach for online connectivity maintenance considering backbones of varying sizes. To the best of our knowledge, this is the first work to tackle the communication-restricted trajectory planning for multi-robot systems considering a manipulator formulation.

\section{Problem formulation}
\label{sec:formulation}

Given a fully-known static planar environment $\mathcal{E}$, a set of $n$ mobile robots $\mathcal{R} = \{r_1, r_2, ..., r_n\}$, and a leader robot $r_l$. The robots work as \textit{routers}, relaying messages over the network between a base station and the leader robot, and they have a communication radius of at most $c_{r, i}$ meters (which can be different for each robot $i$).

Next, we define some concepts used throughout the text. 

\begin{definition}
    \textbf{Communication backbone:} An end-to-end ad hoc wireless network formed by mobile robots able to work as \textit{routers}. The backbone will be used to connect a single leader robot $r_l$ to a fixed based station.
\end{definition}

\begin{definition}
    \textbf{Backbone configuration:} A set of Cartesian coordinates in $\mathbb{R}^2$ that defines the positions of each robot of the backbone, according to a given frame.
    \label{def:backbone_configuration}
\end{definition}

\begin{definition}
    \textbf{Backbone adaptation:} The step of planning a valid trajectory for each robot in the backbone in order to change it from a start configuration to a target one. 
    \label{def:backbone_planning}
\end{definition}

\begin{definition}
    \textbf{Robot connectivity:} Two robots are considered connected if the distance between them is less or equal to a given communication radius $c_r$ and they have direct visibility to each other.
    \label{def:robots_connected}
\end{definition}

\begin{definition}
    \textbf{Valid backbone trajectory}. A backbone trajectory is considered valid if every robot in the backbone is connected to its neighbors during the entire trajectory.
    \label{def:valid_trajectory}
\end{definition}

\begin{problem}[Communication Backbone Reconfiguration with Connectivity Maintenance]
Given a leader robot $r_l$ that can freely move on the environment and a set of mobile routers $\mathcal{R}$. The goal is to sequentially design and deploy a communication backbone that guarantees the leader robot is constantly connected to a fixed base station while it is traversing the environment. The backbone must use a minimal set of routers and minimize the path distance cost during its adaptation.
\end{problem}

\section{Methodology} \label{sec:methodology}

\begin{figure}[htpb]
    \centering
    \includegraphics[width=.85\linewidth]{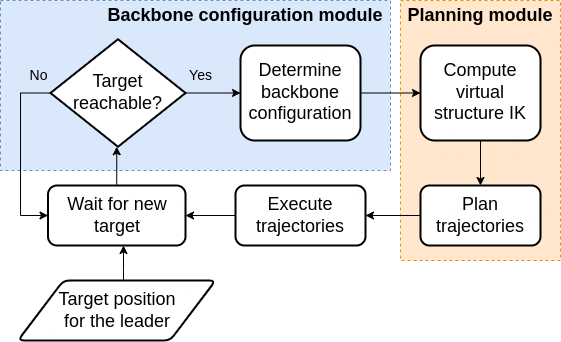}
    \caption{Outline of the full pipeline.}
    \label{fig:system-overview}
\end{figure}

Our method comprises two main components (Fig.~\ref{fig:system-overview}). The first one assesses whether the target is reachable with the available number of relay robots. If so, it then determines the minimum number of relay robots required and their goal positions within the workspace to establish a communication backbone between the base station and the leader robot once it reaches its target position.
According to Def. \ref{def:backbone_configuration}, this is what we call determining the \textit{backbone configuration}. 

The second component is responsible for planning paths for the leader and all robots that are part of the backbone, so that they navigate from their current to target positions keeping the leader connected to the base station at all time. This component receives the goal backbone configuration as input, compute the inverse kinematics (IK) of the virtual structure that corresponds to this configuration, and then plan the paths from the current configurations to the goal configurations. The robots that are not part of the backbone are sent back to the base station (where they can be recharged, for instance).

In a typical mission, these two components run sequentially. The only input is the target position for the leader, and the outputs are the trajectories to each robot, which are then executed using an external trajectory following module. Once all robots navigate through their assigned trajectories and the leader reaches the target, the system waits until a new target position is received and starts the pipeline again.

\subsection{Robot arm modeling} \label{sec:arm_modeling}

We tackle the backbone design and deployment problem as a manipulator planning problem. In other words, we indirectly plan for the robots of the backbone by planning for a virtual structure~\cite{Lewis1997HighPrecision} that represents it. Thus, the first step is to model the robot arm for which the planning will be performed, which depends only on the communication radius and the maximum number of robots in the team.

\begin{figure}[htpb]
    \centering
    \includegraphics[width=0.6\linewidth]{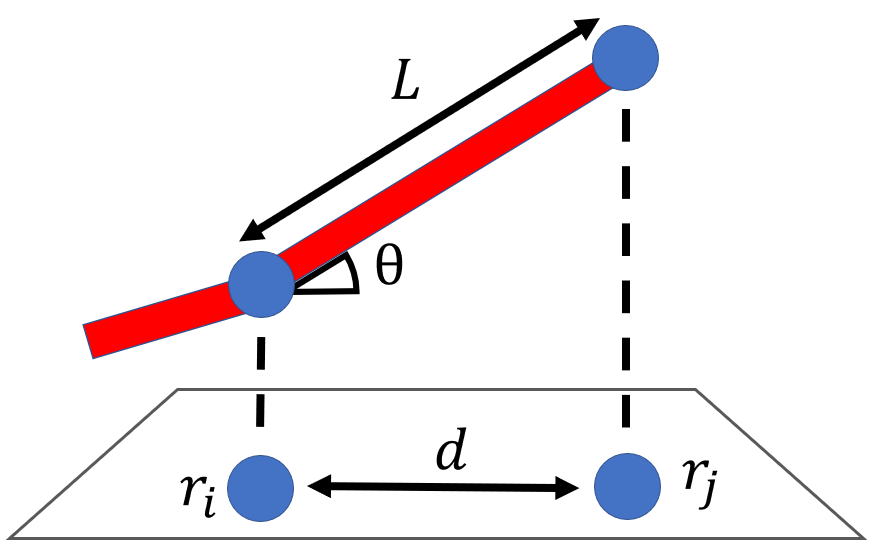}
    \caption{Manipulator draw in perspective as if the $z\text{-axis}$ was parallel to the page to demonstrate that the 3D manipulator approach solves the fixed distance limitation by varying $\theta$.}
    \label{fig:distance_example}
\end{figure}

Our methodology uses a three-dimensional arm modeled as follows. Each robot will be represented by a \emph{universal joint} to allow motion with 2 Degrees of Freedom (DoF), and the base station is mapped to the base link. Hence, each robot in the team must have a fixed index assigned to it from the beginning, which will define its two neighbors and which joint corresponds to it. The link connecting two of these joints represents the wireless connection of the respective robots. Consequently, according to Def.~\ref{def:robots_connected}, we map a loss of connection into a link collision, satisfying the visibility connection constraints while planning. In networked multi-agent path planning, it is allowed - and desirable - that the distances between neighbors vary as long as they respect the communication radius. To that end, the trajectory of each robot is taken as the projection of the corresponding joint on the ground plane along the arm trajectory. Using the orthogonal projection of the joints as the robots' positions allows the distance $d$ between connected robots $r_i$, $r_j$ to vary by changing the pitch angle of the joint $\theta$ while the link length $L$ is kept constant, as shown in Fig.~\ref{fig:distance_example}. The value of $\theta$ is calculated using the backbone configuration as input to an inverse kinematics algorithm described in Sec.~\ref{sec:trajectory_planning}.

To ensure that the distance between neighbor robots is never greater than the communication radius - which would disconnect them - we make $L$ equal to $c_r$. That way, we ensure that neighbor robots are always within the connection radius of each other, satisfying the connection constraint.

\begin{figure}[htpb]
    \centering
    \subfigure[This configuration uses 2 out of 4 robots in the backbone and sends the other two to the base station.]{
      \centering
      \includegraphics[width=0.45\linewidth]{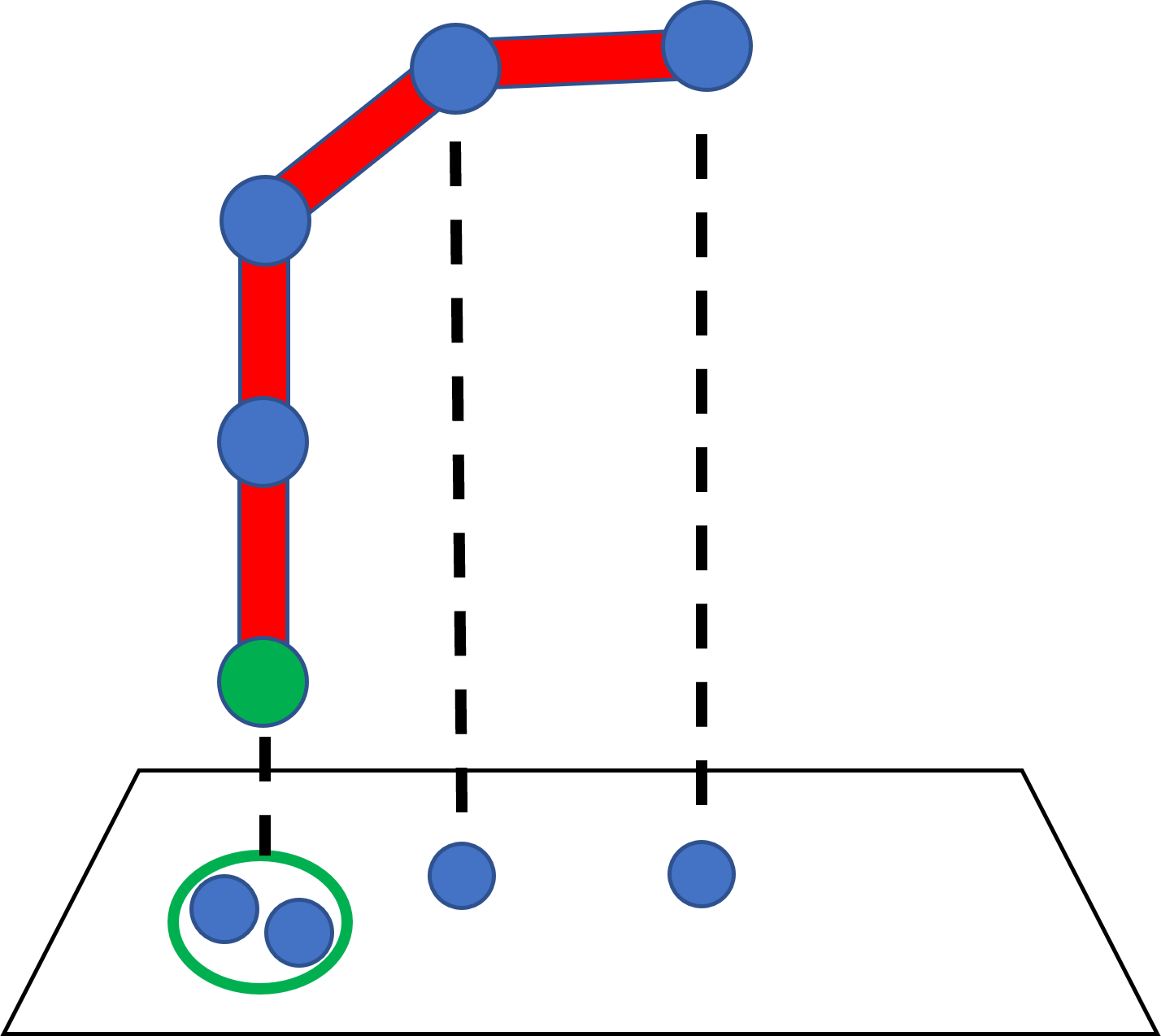}
      \label{fig:diff_number_a}
    }
    \subfigure[This configuration uses 3 out of 4 robots in the backbone and sends the other one to the base station.]{
      \centering
      \includegraphics[width=0.45\linewidth]{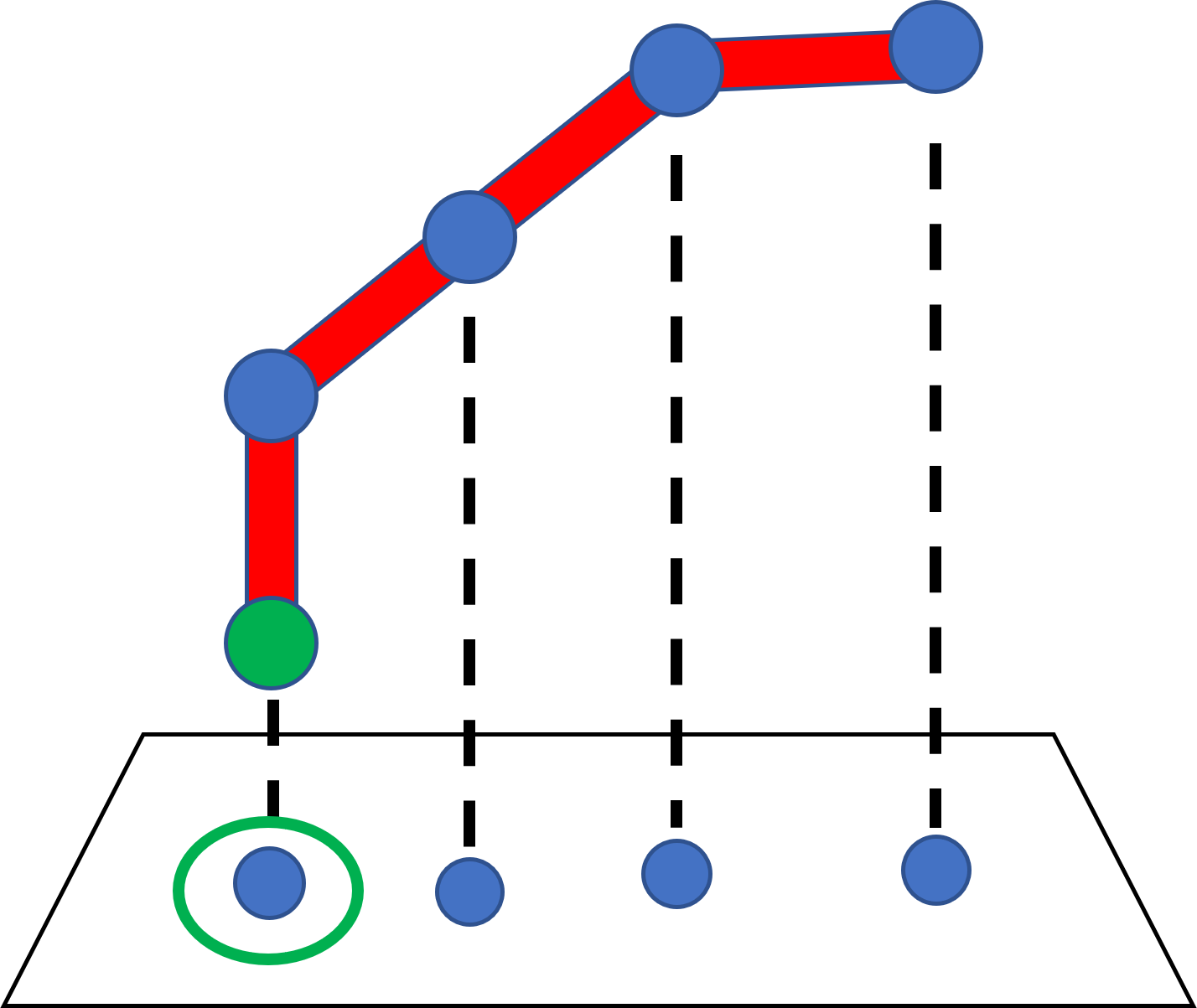}
      \label{fig:diff_number_b}
    }
    \caption{Backbone configurations composed by a different number of robots without changing the robot arm model. The base station is represented by the green base joint, and the robots are represented by the blue joints.}
    \label{fig:diff_number}
\end{figure}

The 3D formulation shown above also allows planning trajectories between backbones composed of a different number of robots. Since we take the orthogonal projection of a position in space as the robots' positions, it is enough to make the $(x, y)$ position of the joints correspondent to the robots that are not needed in the backbone the same as the base's position, which is easily achieved by setting the pitch angle to $0^\circ$ as shown in Fig.~\ref{fig:diff_number}. Note that the backbone configuration may not include all available robots, but the manipulator configuration always does. Besides, by using only universal joints, we are able to cover all valid backbone configurations in the sense that, for all possible backbone configurations, given the network and number of robots parameters, there is an arm configuration in which the projection corresponds to this configuration.

\subsection{Backbone deployment} \label{sec:target_config}

The definition of the target backbone configuration (see Def.~\ref{def:backbone_configuration}) is independent of the planning pipeline. Therefore, it is possible to use external algorithms as long as they respect the visibility constraint, such as~\cite{Santos2014Autonomous}. Nonetheless, we propose a solution for that task in Alg.~\ref{alg:backbone_alg2}, which leans on the visibility network modeling adopted in this work. Function \texttt{DilateObstacles} in Alg.~\ref{alg:backbone_alg2} dilates the obstacles to avoid collisions, and function \texttt{VisGraphPath} calculates the shortest path between two points using a visibility graph, returning an array of nodes containing $(x,y)$ positions.

\begin{algorithm}[htpb]
    \caption{Backbone deployment}
    \label{alg:backbone_alg2}
    \textbf{Input} occupancy map, base station positions, leader's goal
    \textbf{Output} backbone configuration
    \begin{algorithmic}[1]  
    \vspace{1mm}
    \State $\mathcal{M} \gets \texttt{DilateObstacles}(\textrm{occupancy map})$
    \State $G_l \gets \textrm{leader's goal position}$
    \State $R \gets \textrm{base station position}$ \hfill\Comment{Reference point}
    \State $B \gets \{\}$ \hfill\Comment{Backbone array}   
    \State $S_p \gets \texttt{VisGraphPath}(\mathcal{M}, G_l, R)$ \hfill\Comment{Path array}
    \For {$node \textbf{ in } S_p$}
        \If{$||R - node|| > c_R - \delta$}  \Comment{$\delta$ is a constant}
            \State $b_n \gets R + \frac{R - node}{||R - node||}(c_R - \delta)$
            \State $\textrm{Append } b_n \textrm{ to } B$
            \State $R \gets p$
        \Else
            \State $\textrm{Append } node \textrm{ to } B$
            \State $R \gets node$
        \EndIf
  \EndFor
  \State \Return $B$
  \end{algorithmic}
\end{algorithm}

\subsection{Trajectory planning} \label{sec:trajectory_planning}

The main contribution of this paper is the formulation of backbone planning problems as a manipulator planning problem. To plan the backbone trajectory, it is necessary to calculate the arm inverse kinematics corresponding to a given backbone configuration. The arm configuration must be such that each joint's projection on the ground plane corresponds to its associated robot position in the backbone. That way, given a backbone configuration, the planner can calculate the trajectory between the current virtual arm configuration (the current backbone configuration) to the target arm configuration (the target backbone configuration). The inverse kinematics algorithm is given by Alg.~\ref{alg:inv_kin_alg}. Since both positive and negative $z$ would result in the same projected point, we chose to always use the positive $z$-height.

\begin{algorithm}[htpb]
\caption{Inverse kinematics from backbone}
\label{alg:inv_kin_alg}
\begin{algorithmic}[1]
\Require the backbone is connected
\State $B \gets \textrm{backbone configuration}$
\State $N \gets \textrm{team size}$ \Comment{Number of robots available}
\State $c_R \gets \{c_{R, 0}, ..., c_{R, N}\}$ \Comment{Connection radius}
\State $K \gets \{\}$  \Comment{Angles for all (universal) joints}
\State $i \gets 1$  \Comment{Index of the current robot; 0 is the base}
\While{$i \leq N$}
\State $(x_i, y_i) \gets B[i]$
\State Transform $(x_i, y_i)$ to previous joint's frame 
\State $d = ||(x_i, y_i)||$

\If{$d = 0$}
\State $(\theta, \psi) \gets (0, 0)$

\ElsIf{$x_i = 0$}
\State $\theta \gets \arccos{\frac{d}{c_R[i]}}$
\State $\psi \gets \arcsin{\frac{y_i}{d}}$

\Else
\State $\theta \gets \textrm{sign}(x_i)\arccos{\frac{d}{c_R[i]}}$
\State $\psi \gets \textrm{sign}(x_i)\arccos{\frac{y_i}{d}}$
\EndIf

\State Append $(\theta, \psi)$ to $K$
\State $i = i + 1$

\EndWhile

\end{algorithmic}
\end{algorithm}

\begin{figure}[htpb]
    \centering
    \includegraphics[width=0.6\linewidth]{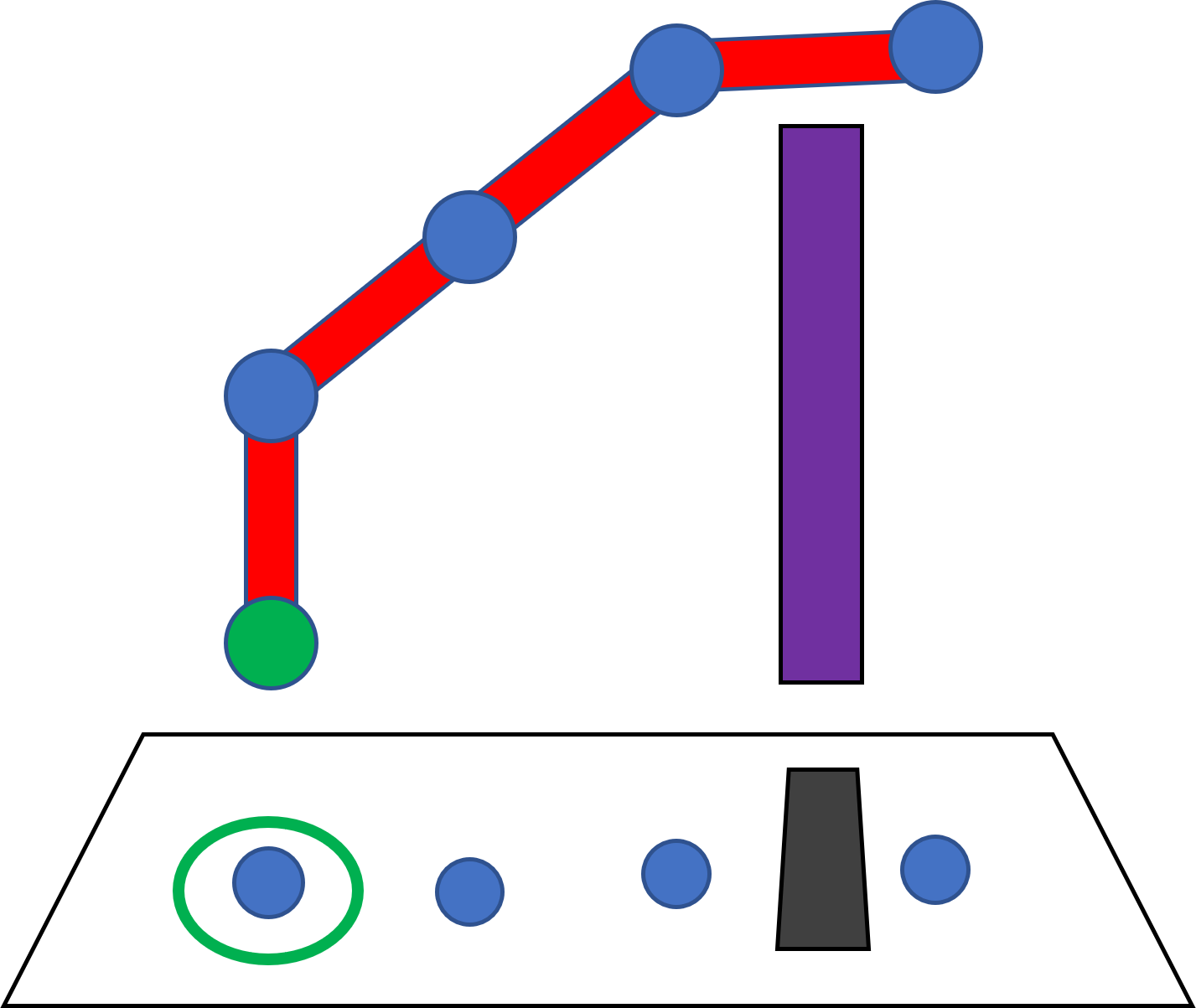}
    \caption{Example in which the manipulator configuration is valid, but the backbone configuration is not, because the arm is above the obstacle. The 3D obstacle is depicted in purple, and the 2D obstacle in gray.}
    \label{fig:3d_obstacle}
\end{figure}

Since the robots are moving in 2D and the planning is being performed in 3D, we must modify the 2D map to avoid valid arm configurations that map to invalid backbone configurations as shown in Fig.~\ref{fig:3d_obstacle}. If $H$ is the arm's height in a full-vertical pose and $\mathcal{C}_{obs, 2D}$ is the set of obstacles in the 2D map, the set of obstacles $\mathcal{C}_{obs, 3D}$ in the 3D map in which planning will be performed is obtained by the Cartesian product $\mathcal{C}_{obs, 3D} = \mathcal{C}_{obs, 2D} \times [0, H]$.

This is a sufficient condition because it guarantees the arm projection will never intersect any obstacle since it will never be able to move above the obstacles.

\begin{figure}[htpb]
    \centering
    \includegraphics[width=.95\linewidth]{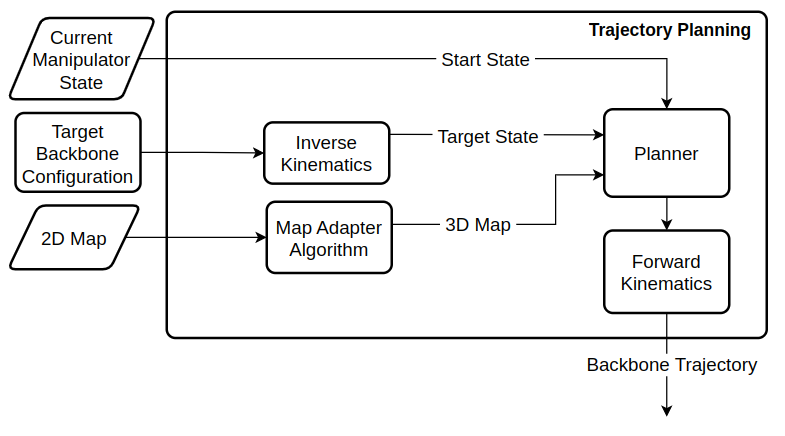}
    \caption{Diagram detailing the trajectory planning process.}
    \label{fig:planning_diagram}
\end{figure}

Once we are able to calculate the arm inverse kinematics from the backbone configuration and modify the 2D map to perform planning in 3D, the backbone planning problem \textit{completely} becomes a manipulator motion planning problem, and we can directly use any planner available in the literature. To avoid infeasible manipulator configurations, we assume all robots start at the base station and compute the initial manipulator pose using Alg.~\ref{alg:inv_kin_alg}. Once the plan is computed for the manipulator, we compute the forward kinematics for every trajectory point and take each joint's $(x, y)$ positions to get the backbone trajectory. Importantly, we are able to reduce planning complexity by planning only for the subset of the joints corresponding to the robots used; \ie, we can disregard the joints corresponding to the robots that remain in the base station during reconfiguration. Optionally, one can use trajectory optimizers to potentially improve the solutions. Note that any solution will be valid according to Def.~\ref{def:valid_trajectory} because the inter-robot distance can never be greater than the virtual manipulator link's length. The planning pipeline is summarized in Fig.~\ref{fig:planning_diagram}.

\section{Experiments} \label{sec:experiments}
\begin{figure*}[htpb]
\centering
    \hspace{-.3cm}
    \subfigure[First backbone transition.]{
        \includegraphics[height=4.1cm]{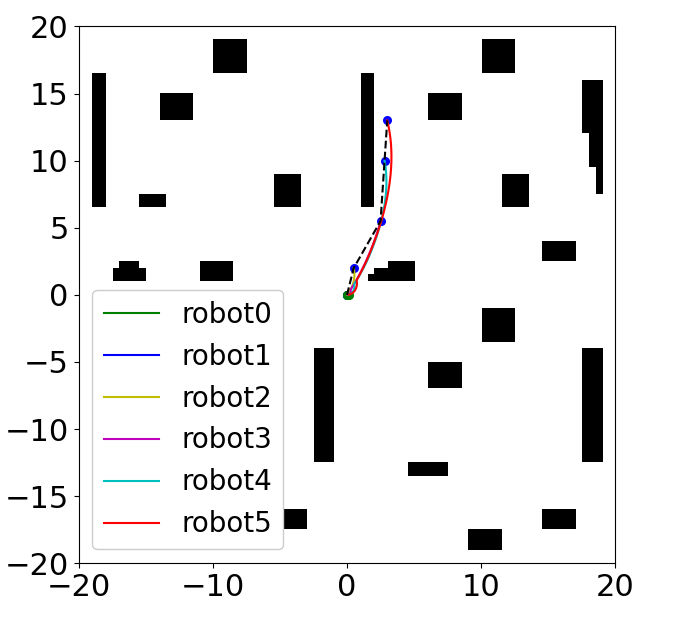}
        \label{fig:illustrative_exp_0}
     }
     \hspace{-.6cm}
     \subfigure[Second backbone transition.]{
        \includegraphics[height=4.1cm]{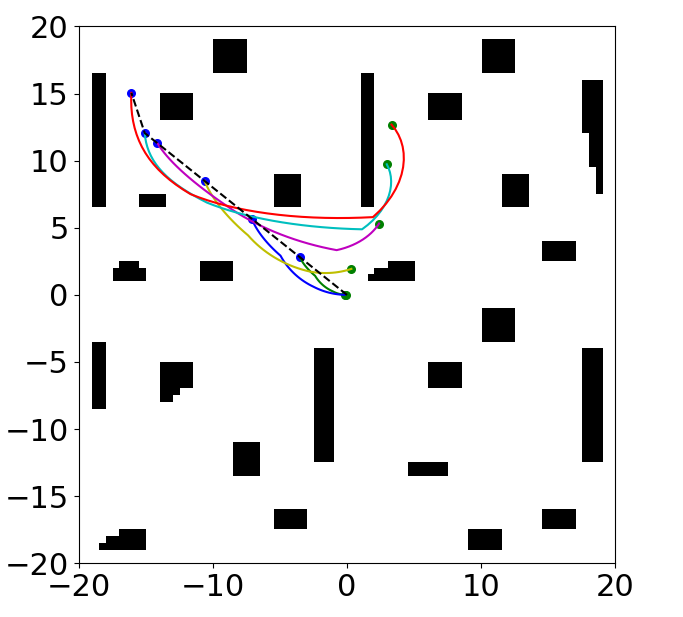}
        \label{fig:illustrative_exp_1}
     }
     \hspace{-.6cm}
     \subfigure[Third backbone transition.]{
        \includegraphics[height=4.1cm]{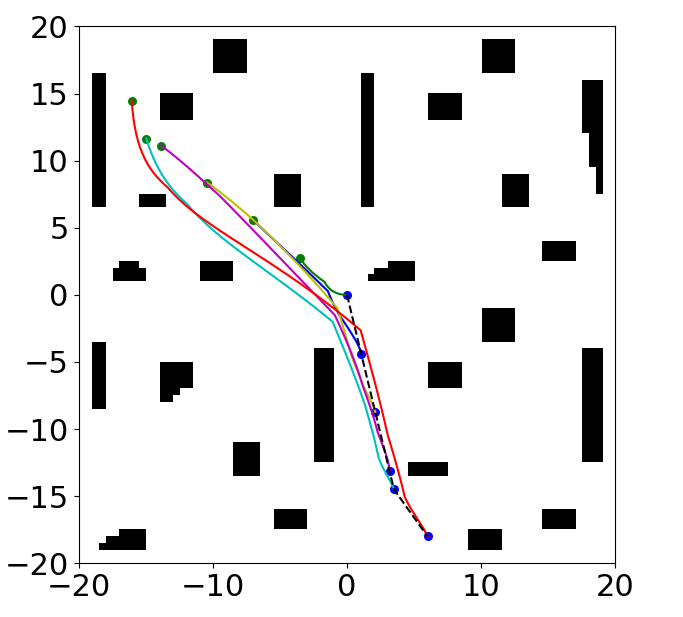}
        \label{fig:illustrative_exp_2}
     }
     \hspace{-.6cm}
     \subfigure[Fourth backbone transition.]{
        \includegraphics[height=4.1cm]{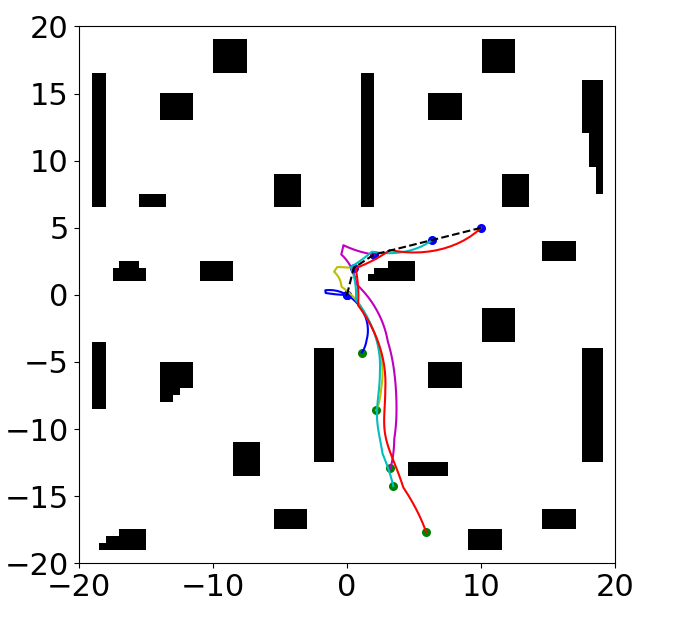}
        \label{fig:illustrative_exp_3}
     }
     \hspace{-.25cm}
     \caption{Example trajectories where \textit{robot5} is the leader. The initial backbone configuration of each sequence is the final backbone configuration of the previous one (see video: \href{https://youtu.be/j_qQrQnGv9Y}{https://youtu.be/j\_qQrQnGv9Y}). The green dots are the initial positions of the robots, and the blue dots are their goals. The backbone is shown as the black dashed line.}
     \label{fig:illustrative_exp}
\end{figure*}

\begin{figure*}[htpb]
\centering
    \hspace{-.7cm}
    \subfigure[First backbone.]{
        \includegraphics[height=3.9cm]{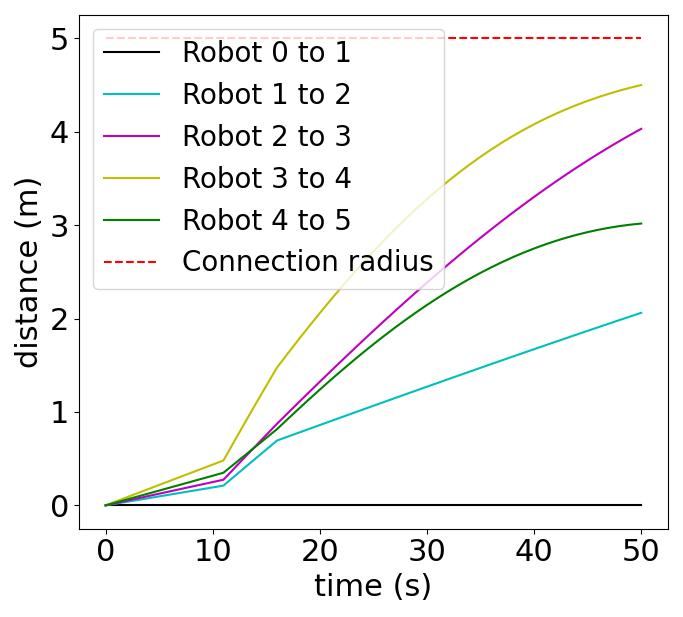}
        \label{fig:illustrative_exp_dists_0}
     }
     \hspace{-.5cm}
     \subfigure[Second backbone.]{
        \includegraphics[height=3.9cm]{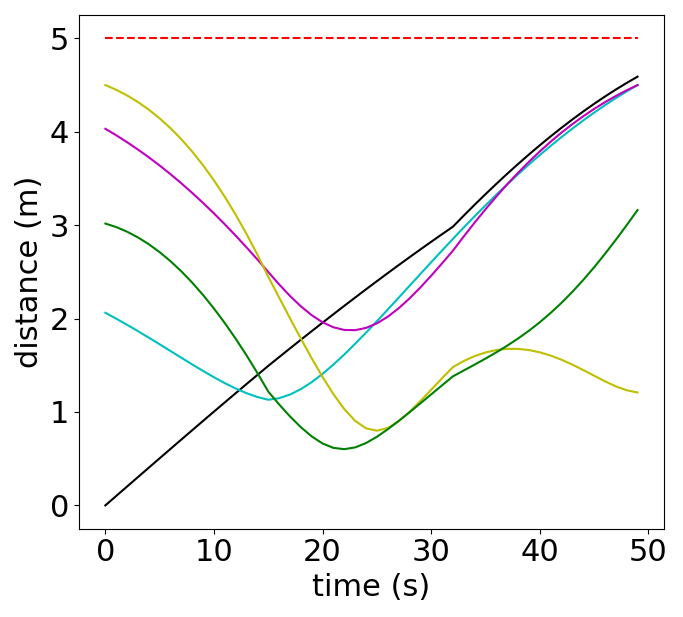}
        \label{fig:illustrative_exp_dists_1}
     }
     \hspace{-.5cm}
     \subfigure[Third backbone.]{
        \includegraphics[height=3.9cm]{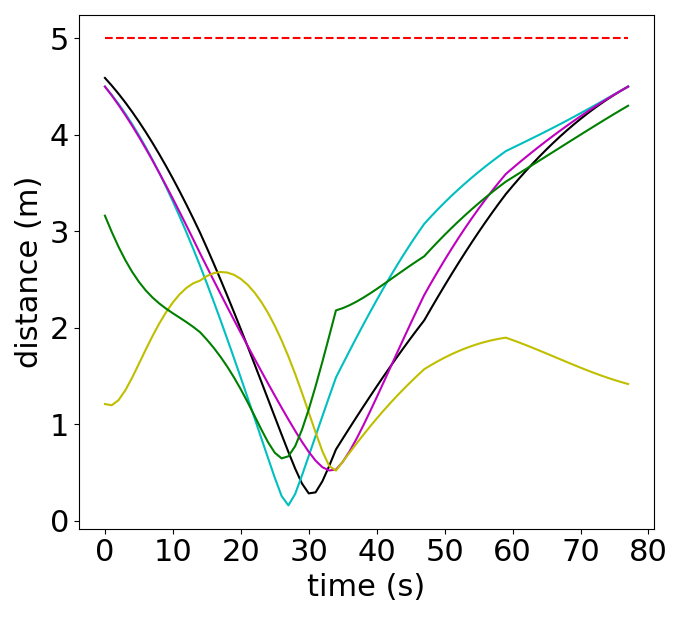}
        \label{fig:illustrative_exp_dists_2}
     }
     \hspace{-.5cm}
     \subfigure[Fourth backbone.]{
        \includegraphics[height=3.9cm]{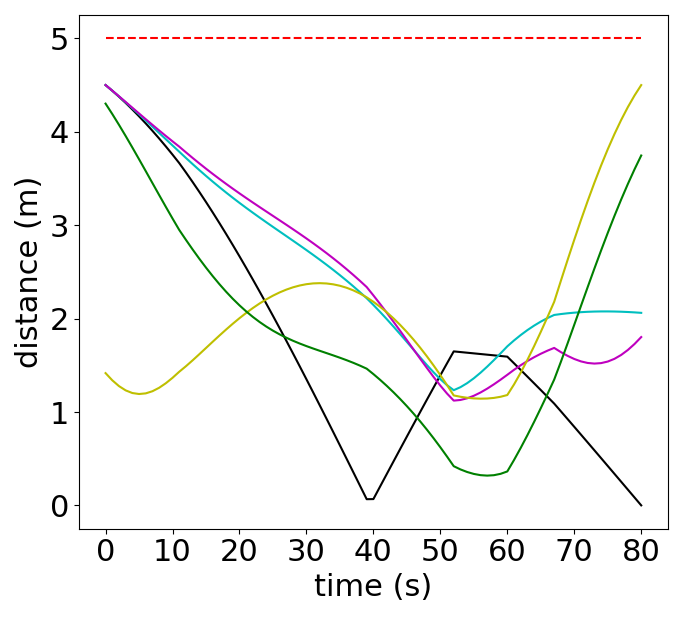}
        \label{fig:illustrative_exp_dists_3}
     }
     \caption{Distance between neighbors along the trajectories. The red dashed line represents the communication radius.}
     \label{fig:illustrative_exp_dists}
\end{figure*}

The simulation framework was implemented in Python 3.8 using ROS Noetic and MoveIt!~\cite{coleman2014moveit}. Simulations were run in Ubuntu 20.04. The planner parameters are shown in Tab.~\ref{tab:illustrative_exp}; and were tuned empirically to achieve a good trade off between planning time and distance cost for multi-robot systems of up to 10 robots.

\begin{table}[htpb]
  \centering
  \caption{Planner parameters for the illustrative example.}
  \label{tab:illustrative_exp}
  \begin{tabular}{ll}
  \toprule
  \textbf{Parameter}    & \textbf{Value}\\
  \midrule
  Planner               & BiTRRT \cite{devaurs2013bitrrt} \\
  Maximum planning time \textit{for one attempt} & $20$s \\
  Maximum attempts within planning time & $200$ \\
  \bottomrule
  \end{tabular}
\end{table}

\subsection{Illustrative example} \label{sec:illus_exp}
Let us first consider an illustrative example to visualize the property of planning between backbones with different numbers of robots and sizes and intuitively evaluate the trajectories. For that, we randomly chose four different positions for the leader using an uniform distribution, each position resulting in a different backbone. All robots started at the base station at the origin. The planned trajectories, shown in Fig.~\ref{fig:illustrative_exp}, are the least costly among a maximum of $200$ solutions computed within the limit of $20$ seconds. If the planner can't find a solution in $20$ seconds, it restarts for another batch of (at most) $200$ attempts in $20$ seconds.

In Fig.~\ref{fig:illustrative_exp_dists} we show that the planned trajectories respect the communication distance, thus maintaining the backbone connected. In this case, we used a safety gap of $0.5$~meters, therefore, the maximum allowed distance between neighbors was $4.5$~meters.
Fig.~\ref{fig:illustrative_exp_dists} also shows that our approach might generate colliding trajectories, which must be addressed by reactive obstacle avoidance in trajectory tracking, such as proposed by Rezende~\etal~\cite{adriano2022vectorfield}.

\subsection{Quantitative analysis}

\begin{figure}[htpb]
    \centering
    \includegraphics[width=.7\linewidth]{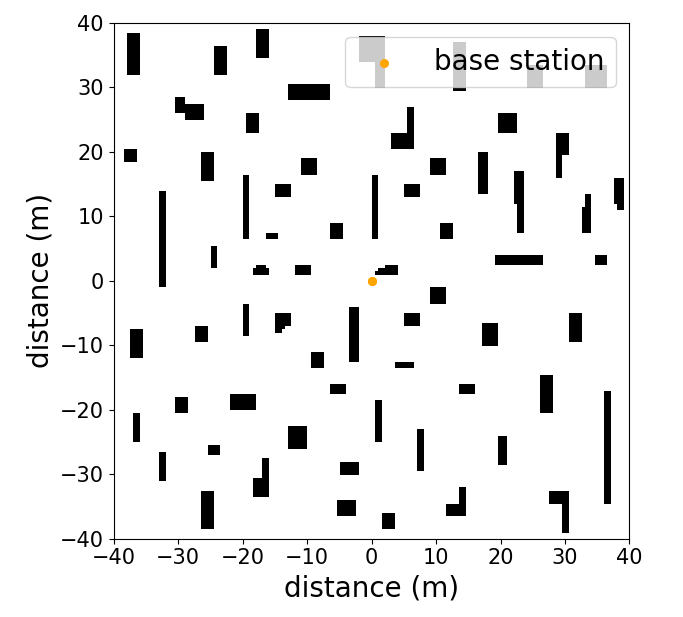}
    \caption{Map used in the quantitative analysis.}
    \label{fig:num_anal_map}
\end{figure}

The map used for the quantitative analysis is shown in Fig.~\ref{fig:num_anal_map}. It is an extended version of the map shown in Fig.~\ref{fig:illustrative_exp} that allows for backbones with more than six robots - given the communication radius of $5$~meters, it would be unlikely to have backbones of $7$ robots or more in a $40$m x $40$m map with the base at the origin.

Similarly to Sec.~\ref{sec:illus_exp}, we randomly chose $10$ positions for the leader using an uniform distribution. To evaluate the impact of the planner's probabilistic nature on the methodology, we calculated trajectories for each position $4$ times.

\begin{figure}[htpb]
    \centering    \includegraphics[width=.8\linewidth]{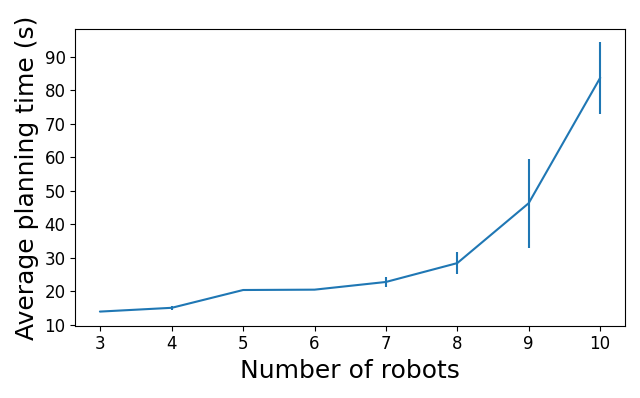}
    \caption{Average planning time over 10 planned trajectories. The error bar shows the standard deviation over 4 trials.}
    \label{fig:avg_plan_time}
\end{figure}

Fig.~\ref{fig:avg_plan_time} shows the average time needed to plan the trajectories for different team sizes. Because our approach is centralized, it does not scale well with the number of robots. However, as we are directly applying the planning algorithm with no modifications, we still preserve the asymptotic optimality. This is a common and expected trade-off between centralized and decentralized approaches.

\begin{figure}[htpb]
    \centering
    \includegraphics[width=.8\linewidth]{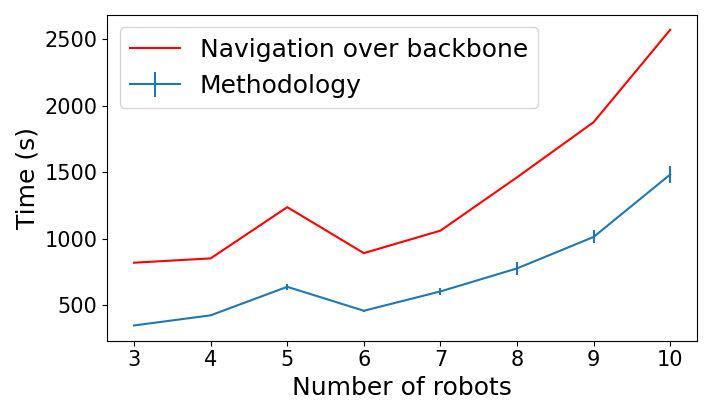}
    \caption{Mission execution time using the simple, but rigid, navigation over the backbone versus our methodology. The error bar shows the deviation over 4 trials.}
    \label{fig:time_comparison}
\end{figure}

Finally, since we argue simplicity is a major advantage of our methodology, we compare the performance to an equivalently simple approach used in \cite{Santos2019Fast}, which is to navigate the robots over the backbone. A drawback of this approach is that the team cannot move simultaneously because they lose communication if a robot does not wait for its neighbors in one of the nodes of the backbone. As Fig.~\ref{fig:time_comparison} shows, because our method allows the robots to move simultaneously, we are able to finish the mission (visit all positions assigned to the leader) 80\% faster for teams of 10 robots - a critical advantage in disaster response operations. We assumed the robots navigate at a maximum velocity of $0.5$ m/s.

\subsection{Hardware experiment}
\begin{figure*}[htpb] 
\centering
	\subfigure[$t=0$s]{
	    \includegraphics[width=.24\linewidth, trim=200 20 425 40,clip]{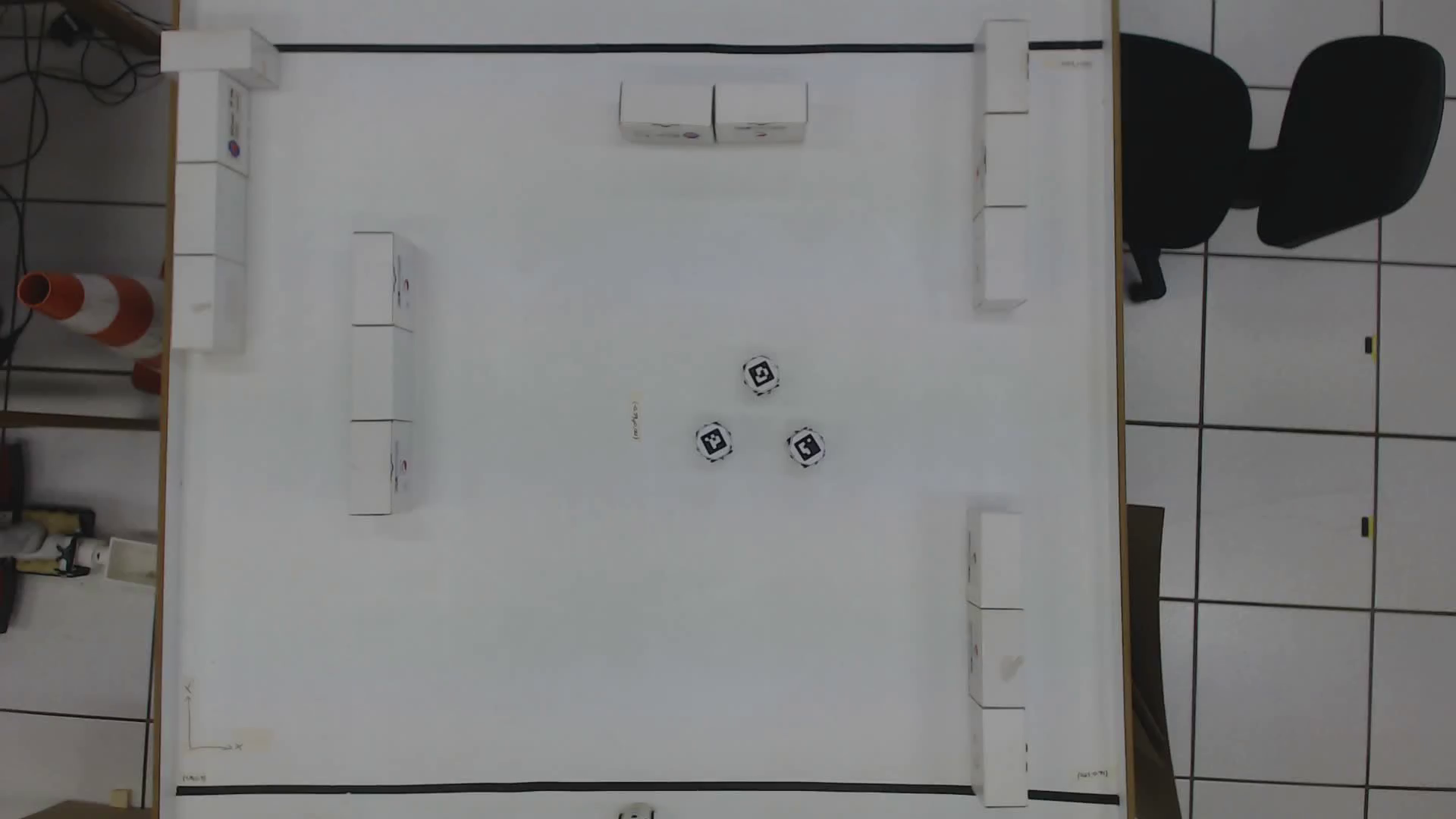}
	}
	\hspace{-.4cm}
	\subfigure[$t=66$s]{
	    \includegraphics[width=.24\linewidth, trim=200 20 425 40,clip]{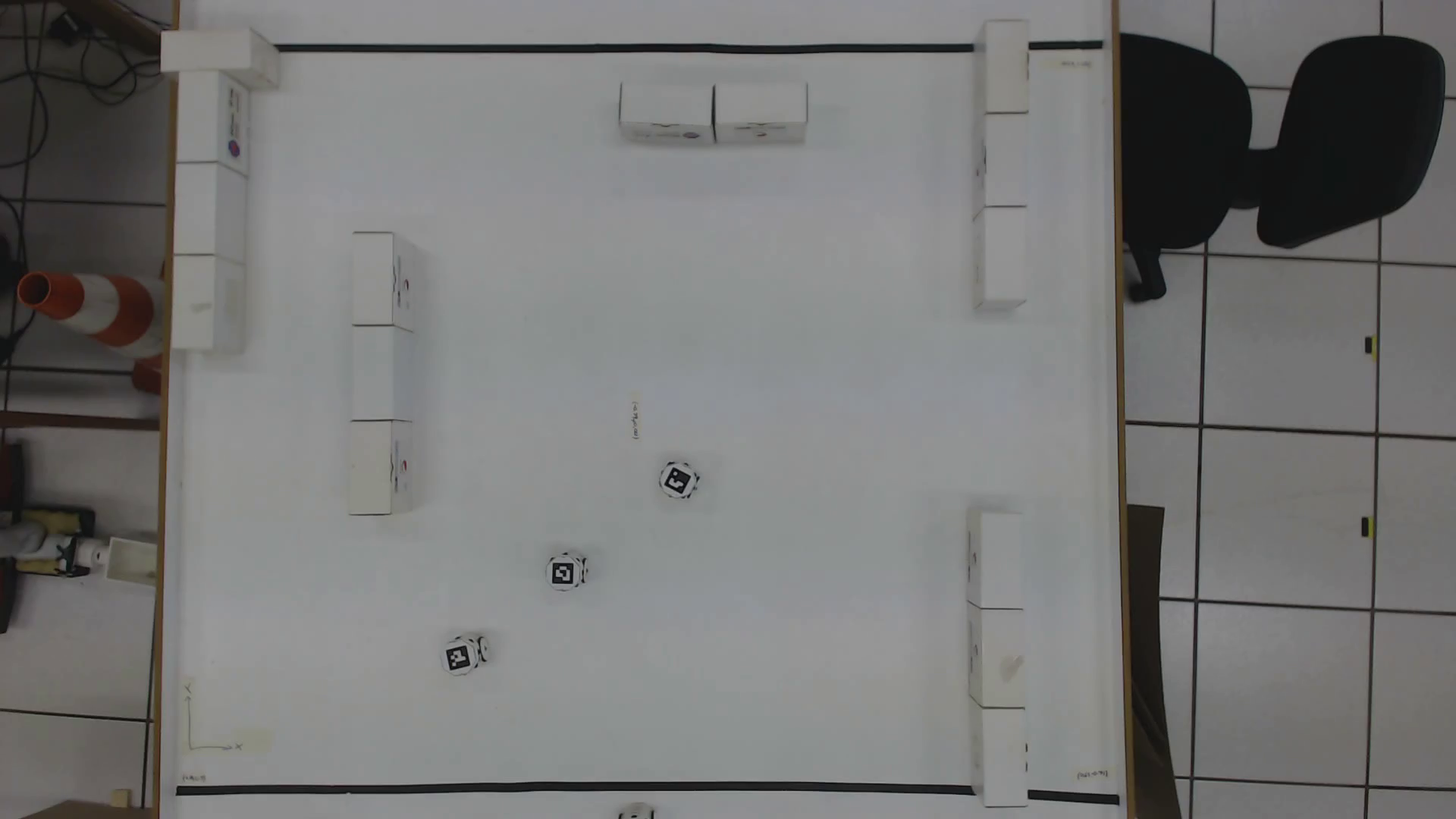}
	}
	\hspace{-.4cm}
	\subfigure[$t=105$s]{
	    \includegraphics[width=.24\linewidth, trim=200 20 425 40,clip]{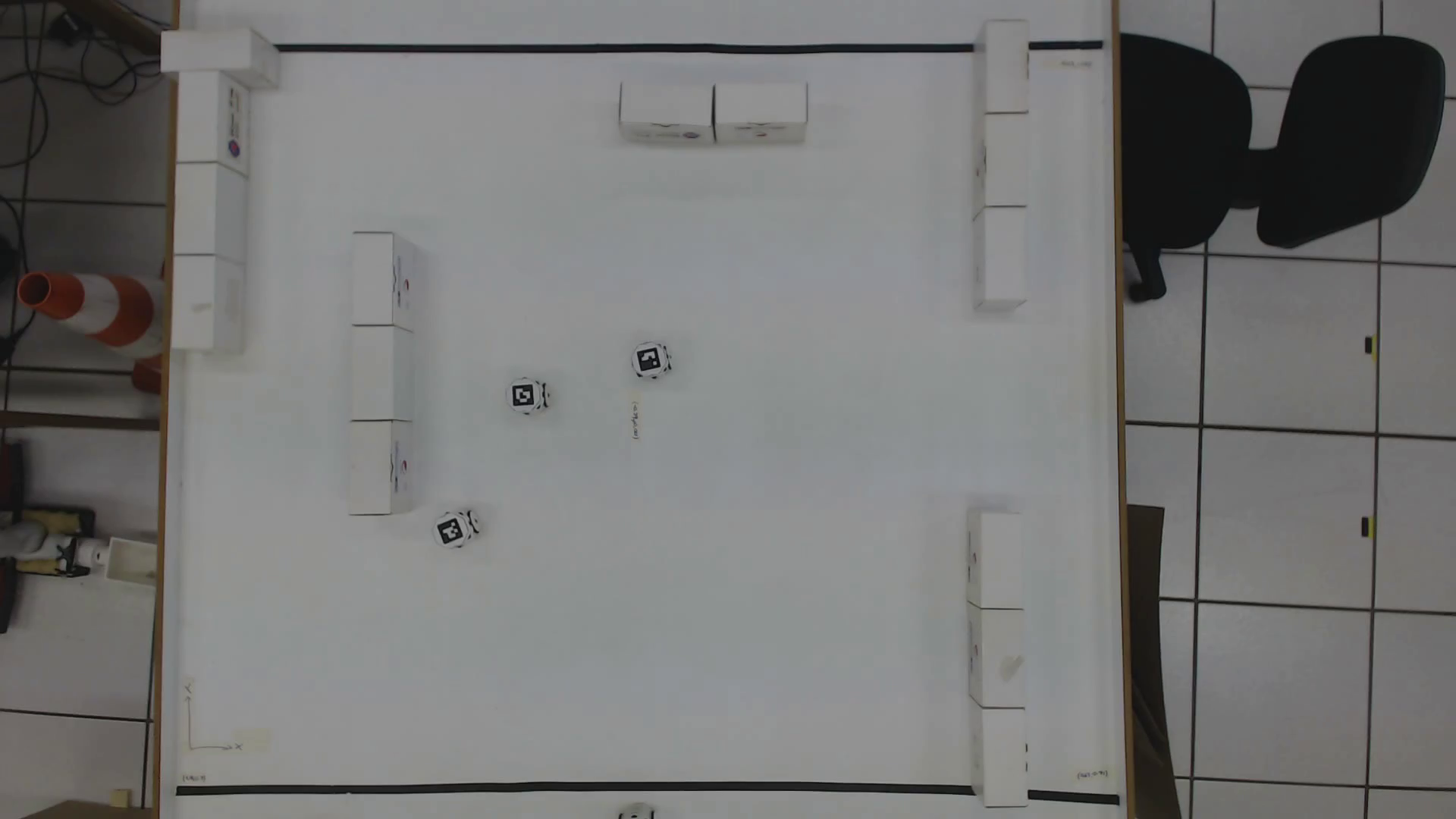}
	}
	\hspace{-.4cm}
	\subfigure[$t=150$s]{
	    \includegraphics[width=.24\linewidth, trim=200 20 425 40,clip]{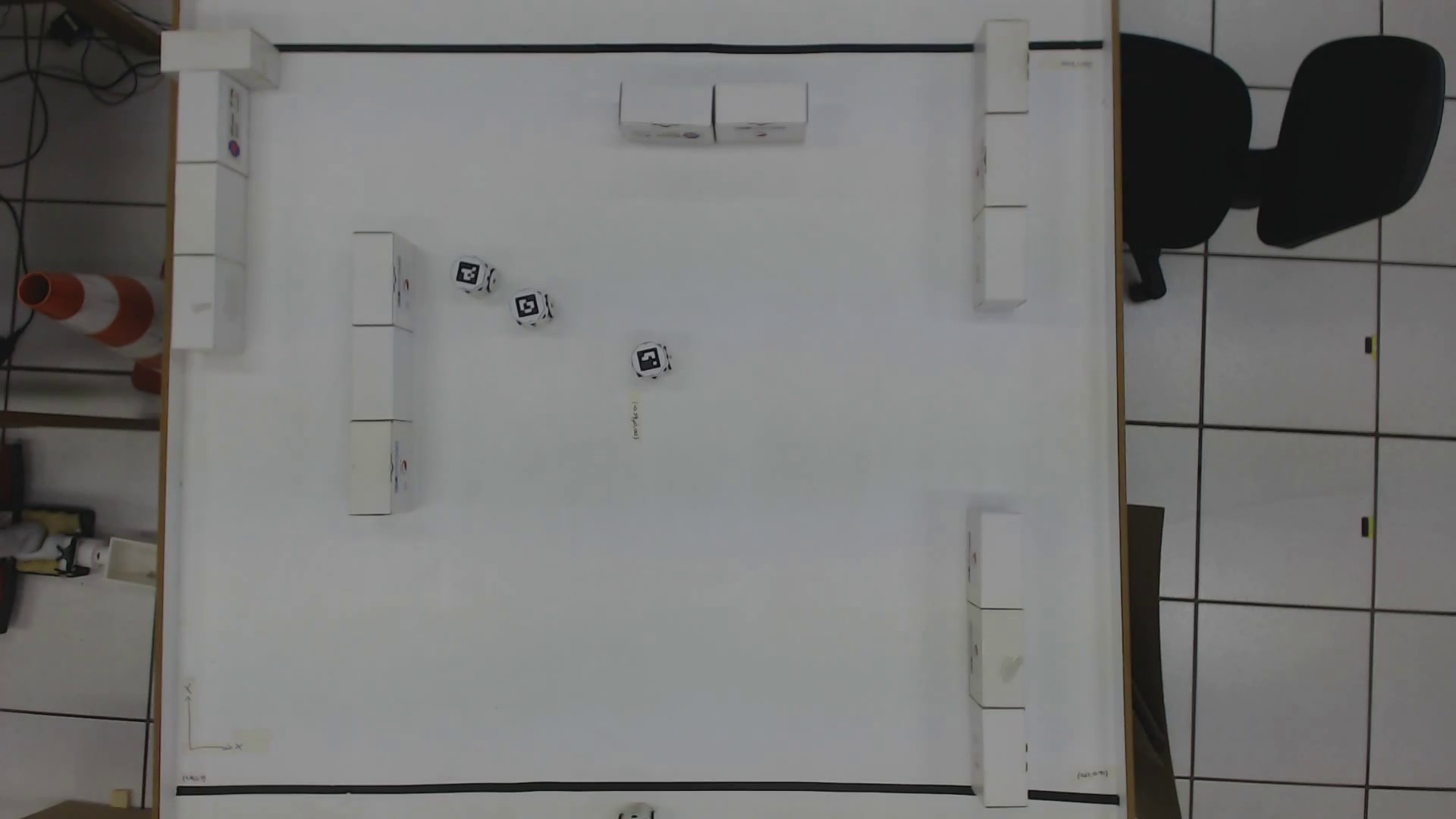}
	}
	\caption{Real-world proof-of-concept execution in an obstacle-free environment. Video: \href{https://youtu.be/Yj-UzMRdQbI}{https://youtu.be/Yj-UzMRdQbI}}
	\label{fig:experiment_real}
\end{figure*}

We also evaluated the approach with a real robot deployment in an obstacle-free scenario. Snapshots of the execution are shown in Fig.~\ref{fig:experiment_real} and a video is available online\footnote{\href{https://youtu.be/Yj-UzMRdQbI}{https://youtu.be/Yj-UzMRdQbI}}.

The robotic platform was developed in-house for experiments with swarm robotics~\cite{Rezeck2017HeRo}, and we used a top-view camera for global localization with a tag detection method.

The scenario has $1.78 \text{m} \times 1.40 \text{m}$, and the network radius was set as $30 \text{ cm}$. Each robot was controlled individually using the the kinematic controller proposed in \cite{desai1998controlling} and deviated from the other robots in a reactive way.

As Fig. \ref{fig:experiment_real} shows, the robots started at the base, then, established a communication backbone for the leader at $p_1 = (-0.6, -0.5)$. Then, a new backbone at $p_2 = (-0.6, 0.3)$ was required, and the system accompanied the leader keeping it connected to the base.

\section{Conclusion and Future Work} \label{sec:conclusion}

In this paper, we tackled the problem of planning paths for mobile multi-robot systems with connectivity maintenance requirements. We reformulated this problem as a manipulator planning problem and showed that we could use techniques already consolidated in that field for planning problems involving linearly networked MRS. We validated our proposal in different scenarios with small and medium-sized teams. Due to its centralized nature, however, the methodology's performance tends to degrade as the size of the team increases.

In future work, we intend to extend these ideas to design an online planner, where the backbone must provide communication to a moving client (leader robot) considering the end-effector is constantly moving and to include kinematics restrictions during planning to avoid colliding trajectories. Additionally, we also plan to evaluate the use of heterogeneous teams, \eg, with aerial and ground vehicles.

\bibliographystyle{ieeetr}  

\bibliography{bibliography}

\ifCLASSOPTIONcaptionsoff
  \newpage
\fi

\end{document}